\documentclass[10pt,twocolumn,letterpaper]{article}

\usepackage{cvpr}              
\usepackage{gensymb}
\usepackage{multirow}
\usepackage{algorithm}
\usepackage{algorithmicx}  
\usepackage{algpseudocode} 
\usepackage{amsmath}
\usepackage{amssymb}
\usepackage{booktabs}
\usepackage[table]{xcolor}
\usepackage{makecell}
\usepackage{marvosym}
\definecolor{cvprblue}{rgb}{0.21,0.49,0.74}

\definecolor{GammaColor}{rgb}{0.5,0,0.7}
\definecolor{myheadercolor}{RGB}{230, 230, 230} 
\definecolor{mysectioncolor}{RGB}{245, 245, 245} 
\definecolor{myourscolor}{RGB}{215, 246, 255}    
\newcommand{\authorskip}{\hspace{1mm}}

\usepackage[pagebackref,breaklinks,colorlinks,allcolors=cvprblue]{hyperref}

\def\paperID{} 
\def\confName{CVPR}
\def\confYear{2026}

\title{From Pairs to Sequences: Track-Aware Policy Gradients for Keypoint Detection}

\author{
    Yepeng Liu\textsuperscript{1,2 *}, \authorskip 
    Hao Li\textsuperscript{2 *}, \authorskip 
    Liwen Yang\textsuperscript{1 *}, \authorskip
    Fangzhen Li\textsuperscript{2 $\dagger$},\authorskip
    Xudi Ge\textsuperscript{1},\authorskip
    Yuliang Gu\textsuperscript{1},\\
    Kuang Gao\textsuperscript{2},\authorskip
    Bing Wang\textsuperscript{2},\authorskip
    Guang Chen\textsuperscript{2},\authorskip
    Hangjun Ye\textsuperscript{2},\authorskip
    Yongchao Xu\textsuperscript{1 $\dagger$ \Letter}\authorskip\\
\textsuperscript{1}School of Computer Science, Wuhan University \hspace{2mm}
\textsuperscript{2}Xiaomi EV\\
{\tt\small \{yepeng.liu@,yongchao.xu@\}whu.edu, }
{\tt\small \{lifangzhen@\}xiaomi.com}
}

\begin{document}
\maketitle
\let\thefootnote\relax
\footnotetext{
\small
\textsuperscript{*} Equal contribution, 
\textsuperscript{$\dagger$} Project leader,
\textsuperscript{\Letter} Corresponding author.
}

\begin{abstract}
Keypoint-based matching is a fundamental component of modern 3D vision systems, such as Structure-from-Motion (SfM) and SLAM. 
Most existing learning-based methods are trained on image pairs, a paradigm that fails to explicitly optimize for the long-term trackability of keypoints across sequences under challenging viewpoint and illumination changes. 
In this paper, we reframe keypoint detection as a sequential decision-making problem. We introduce \textbf{TraqPoint}, a novel, end-to-end Reinforcement Learning (RL) framework designed to optimize the \textbf{Tra}ck-\textbf{q}uality (Traq) of keypoints directly on image sequences. 
Our core innovation is a track-aware reward mechanism that jointly encourages the consistency and distinctiveness of keypoints across multiple views, guided by a policy gradient method. Extensive evaluations on sparse matching benchmarks, including relative pose estimation and 3D reconstruction, demonstrate that TraqPoint significantly outperforms some state-of-the-art (SOTA) keypoint detection and description methods.
The code will be available at \url{https://github.com/xiaomi-research/traqpoint}.
\vspace{-0.5 em}
\end{abstract}

\section{Introduction}
\vspace{-2pt}
Keypoint detection and description are the foundation for a wide range of 3D computer vision tasks, including Structure-from-Motion (SfM)~\cite{schonberger2016structure, liu2025liftfeat}, Simultaneous Localization and Mapping (SLAM)~\cite{mur2017orb, chung2023orbeez}, and  relocalization~\cite{sarlin2019coarse,  yin2023isimloc}. 
The performance of these systems is critically dependent on the quality of the detected keypoints. To enable robust matching, these points must exhibit high stability under drastic viewpoint and illumination changes.

\begin{figure}[!t]
    \centering
\includegraphics[width=0.48\textwidth]{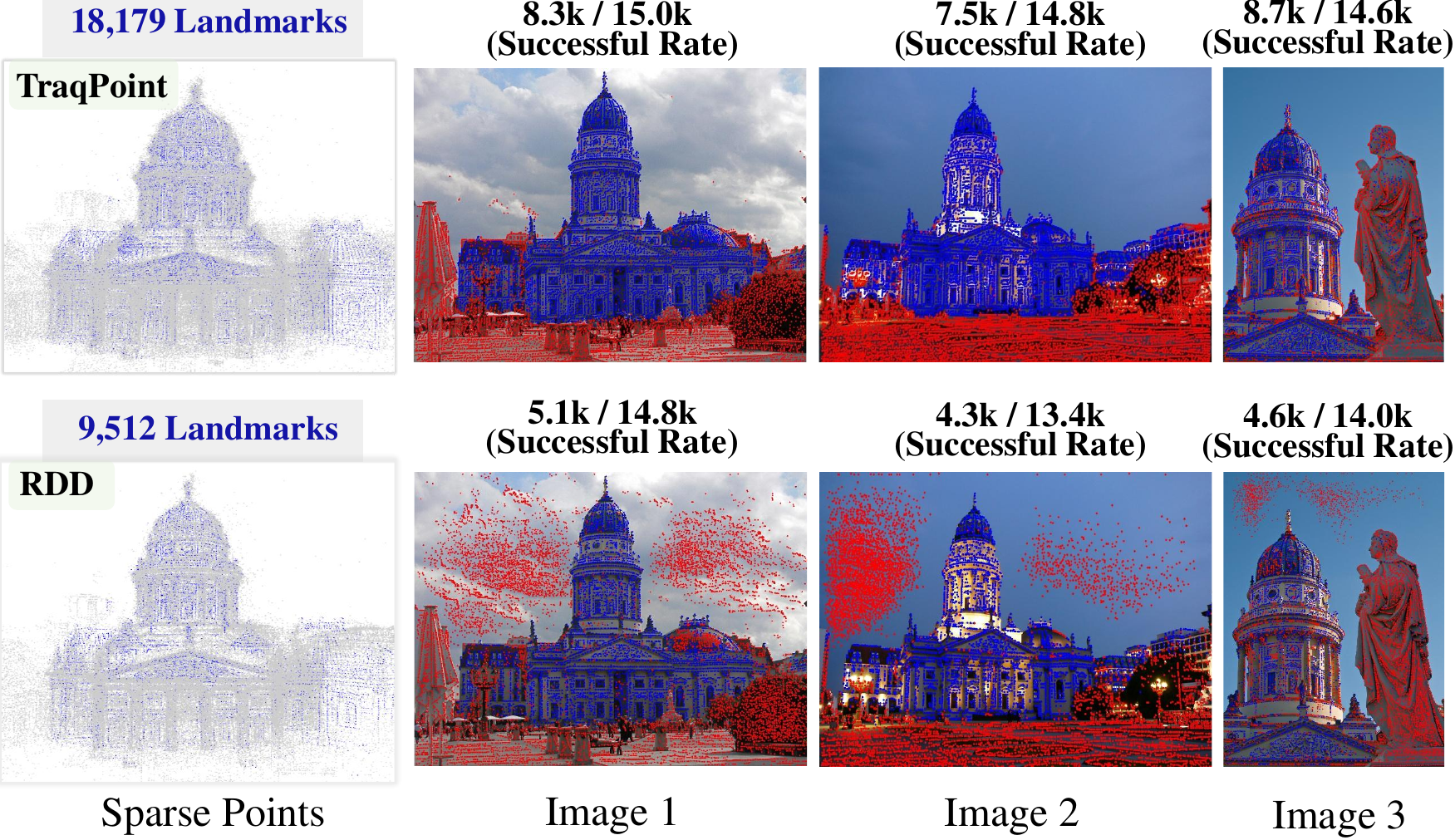}
    \caption{
Multi-view reconstruction: Our TraqPoint vs. RDD~\cite{chen2025rdd}. Keypoints that successfully generate  landmarks are marked in \textcolor{blue}{blue}. Failed keypoints are marked in \textcolor{red}{red}. 
Our TraqPoint generates more keypoints in structurally significant areas with higher cross-view consistency, yielding more landmarks.
}
\label{fig:motivation}
\vspace{-0.8 em}
\end{figure}

Driven by advances in deep learning, the paradigm for keypoint detection has progressively shifted from handcrafted designs to data-driven approaches. These learning-based methods are generally self-supervised or supervised. Self-supervised methods, such as SuperPoint~\cite{detone2018superpoint} and SiLK~\cite{gleize2023silk}, generate pseudo-labels from synthetic image pairs using geometric transformations. 
Supervised approaches \cite{zhao2023aliked, wang2022featurebooster, edstedt2024dedodev2,chen2025rdd} instead train on large-scale real-world datasets (\eg, MegaDepth~\cite{li2018megadepth}) to learn repeatable keypoints.
Another category of methods (RFP~\cite{bhowmik2020reinforced}, DISK~\cite{tyszkiewicz2020disk}, and RIPE~\cite{kunzel2025ripe}) proposes using Reinforcement Learning (RL) to select keypoints in a pair of images, rewarding those keypoints that can form matching pairs.
Despite their varied training strategies, existing methods converge on a common principle: defining keypoint quality through its repeatability and matchability in an image pair.

This pairwise training paradigm has significantly improved performance on matching tasks. However, it inherently optimizes for instantaneous ``matchability" within isolated image pairs. This objective is misaligned with sequential applications (\eg, SLAM and SfM), where the core requirement is long-term ``trackability".
Specifically, keypoints that perform well on a single pair may drift or drop over  long trajectories due to strong viewpoint/illumination changes or motion blur. Such failures directly compromise the stability and accuracy of the overall system.

In this paper, we advocate for a paradigm shift: moving from optimizing ``pairwise matchability" to directly learning for ``long-term trackability". To this end, we introduce a novel \textbf{Sequence-Aware Keypoint Policy Learning} framework. We reframe keypoint detection as a sequential decision-making problem, where the goal is to discover points that yield stable and persistent tracks. In our framework, a policy network acts as a RL agent that operates on a single reference image to select a sparse set of candidate keypoints. The ``environment" is no longer a single paired image but an entire sequence. Correspondingly, the reward is a function of the quality of the entire track generated by each selected keypoint.

Specifically, the reward for a track is a composite of two signals: (1) a \textbf{Rank Reward} that incentivizes selecting points maintaining high saliency within their local neighborhood across multiple views, and (2) a \textbf {Distinctiveness Reward} that encourages selecting points with global uniqueness. By optimizing this track-centric objective, our policy network learns to produce a detection probability map that intrinsically favors points with the highest potential for long-term stability. 
As shown in Fig.~\ref{fig:motivation}, TraqPoint's keypoints are concentrated in structurally significant areas and exhibit superior cross-view consistency compared to the state-of-the-art approach, RDD~\cite{chen2025rdd}.
Extensive experiments confirm that TraqPoint achieves SOTA results—specifically on both pairwise and sequential downstream tasks, including relative pose estimation, visual localization, visual odometry, and 3D reconstruction.

Our main contributions can be summarized as follows:

\begin{itemize}
    \item We  identify a critical gap between the pairwise training paradigm and the demands of sequential applications. To bridge this gap, we pivot to a new perspective and propose a novel Reinforcement Learning (RL) framework that directly optimizes long-term keypoint trackability.
    \item For the RL framework, we propose a hybrid sampling strategy that enables efficient candidate keypoint selection, while also introducing a novel composite reward function to jointly optimize for multi-view  consistency and distinctiveness.
    \item Through comprehensive experiments, we show that TraqPoint significantly outperforms state-of-the-art methods on both pairwise tasks and sequential tasks.
\end{itemize}
\section{Related Work}
\vspace{-0.2 em}
\subsection{Learned Keypoint Detection and Description}
\vspace{-0.2 em}
The definition of a ``good" keypoint has evolved from classical handcrafted~\cite{lowe2004distinctive, matas2004robust, rosten2008faster, rublee2011orb, xu2014tree} designs to modern learning-based paradigms~\cite{savinov2017quad, revaud2019r2d2, lee2022self, pakulev2023ness, kim2024learning, barbarani2024scale, santellani2024gmm}. Traditional algorithms defined core properties of desirable keypoints: SIFT~\cite{lowe2004distinctive} identifies scale-space extrema via Difference-of-Gaussians, valuing scale stability and rotation invariance; FAST~\cite{rosten2008faster} and ORB~\cite{rublee2011orb} prioritize efficiency, detecting corners via surrounding pixel intensities. 
TBMR~\cite{xu2014tree} leverages Morse theory for keypoint extraction, emphasizing contrast and affine invariance.
These methods share the principle that ideal keypoints lie in textured regions with strong, repeatable geometric  signatures (\eg, corners, blobs).

The learning-based paradigm has dramatically improved the robustness of keypoint detection and matching in complex scenes through a data-driven, pairwise training approach. One line of work uses self-supervision, typically by leveraging homography transformations to generate image pairs with known correspondences. SuperPoint~\cite{detone2018superpoint}
, for example, uses a base detector to create pseudo-labels on warped images, enabling joint training of the detector and descriptor. Similarly, R2D2~\cite{revaud2019r2d2} 
optimizes for both repeatability and reliability in a self-supervised manner, while SILK~\cite{gleize2023silk} 
uses the matchability of learned descriptors to supervise which points should be detected. Another category of methods, including D2-Net~\cite{dusmanu2019d2}, ALIKED~\cite{zhao2023aliked}, XFeat~\cite{potje2024xfeat}, DeDoDe~\cite{edstedt2024dedodev2}, and RDD~\cite{chen2025rdd}, trains on large-scale 3D datasets like MegaDepth~\cite{li2018megadepth}. By using ground-truth correspondences derived from COLMAP~\cite{schonberger2016structure} reconstructions, these methods learn features that are highly consistent across real-world viewpoint and illumination changes. 
Furthermore, dense (RoMa~\cite{edstedt2024roma}, Mast3r~\cite{leroy2024grounding}, Dust3r~\cite{wang2024dust3r}) and semi-dense (LoFTR~\cite{sun2021loftr}, Efficient LoFTR~\cite{wang2024efficient}) matching methods have garnered significant attention recently, and they also adopt a pairwise training paradigm.

While these methods have established a strong baseline for pairwise matching, their objective is not explicitly designed to model long-term temporal dynamics. Our work addresses this gap by introducing a sequence-aware paradigm that optimizes directly for long-term trackability. This approach offers greater potential to handle complex factors inherent in sequences.

\subsection{RL-Based Keypoint Detection}
Several works frame keypoint detection as a Reinforcement Learning (RL) problem to handle the discrete and non-differentiable keypoint selection task. Early approaches, such as~\cite{truong2019glampoints, cieslewski2019sips}, utilize Q-learning to directly regress a keypoint's reward value. However, these rewards are calculated based on the matchability of handcrafted descriptors (\eg, SIFT), which fundamentally limits the robustness of the learned detector.

More recent methods~\cite{bhowmik2020reinforced, tyszkiewicz2020disk, potje2023enhancing, santellani2023s, kunzel2025ripe, edstedt2025dad} perform gradient optimization directly in policy space. Reinforced Feature Points (RFP)~\cite{bhowmik2020reinforced} represents a key advancement by embedding the learning process into high-level tasks, using the final pose estimation error as the reward signal to align keypoint selection with 3D geometry. DISK~\cite{tyszkiewicz2020disk} achieves end-to-end optimization by using the number of geometrically consistent inliers as a reward, while RIPE~\cite{kunzel2025ripe} proposes an unsupervised approach that uses the inlier ratio from epipolar constraints. These methods still retain the  limitation of the pairwise paradigm: their reward functions are derived from a single image pair, optimizing for {instantaneous matchability} rather than the {longevity of keypoints} required in sequential tasks.

Our work departs from these RL-based methods by shifting the optimization paradigm from pairwise to sequence-level supervision, computing rewards based on the {trajectory quality} of selected keypoints to directly optimize for ``trackability" over ``matchability".  Furthermore, we decouple policy learning from descriptor training, leveraging a frozen descriptors branch to provide stable reward signal.

\section{Method}
\label{methods}
\begin{figure}[!t]
    \centering
\includegraphics[width=0.48\textwidth]{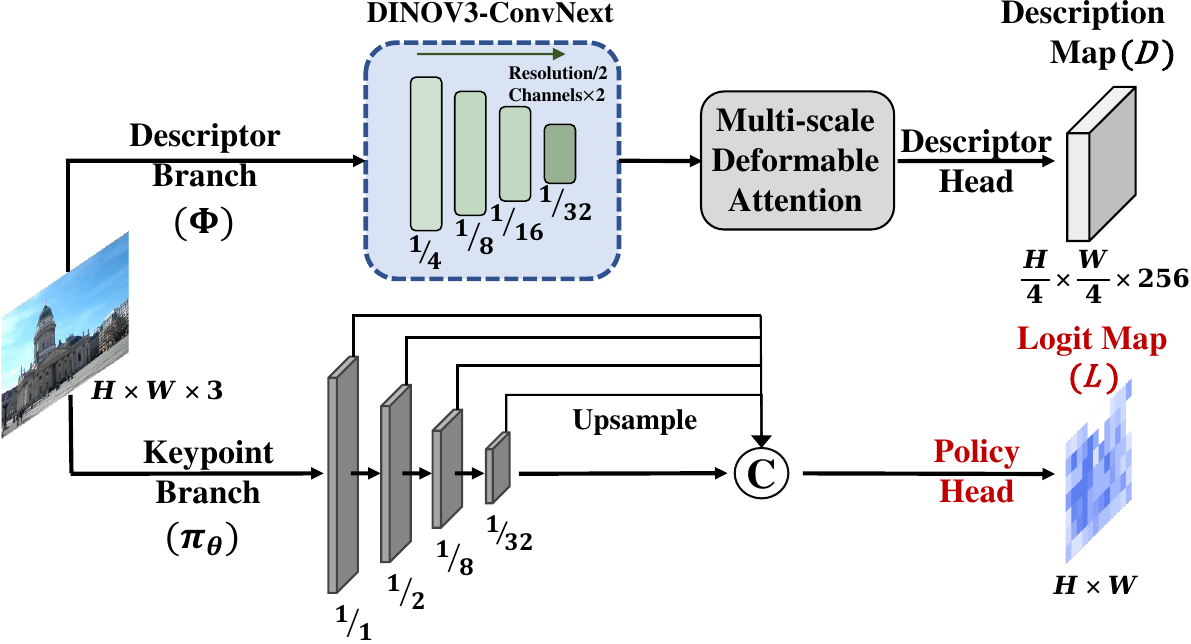}
    \caption{
 Following the architectural design of RDD~\cite{chen2025rdd}, we adopt an identical network structure. Specifically, we replace the feature extractor employed in RDD~\cite{chen2025rdd} with DINOv3-ConvNeXt~\cite{simeoni2025dinov3}.
 The keypoint branch serves as the ~\textbf{policy network ($\pi_\theta$)}.
 }
\label{fig:network}
\vspace{-1 em}
\end{figure}

\begin{figure*}
\vspace{-0.8 em}
    \centering
    \includegraphics[width=0.9\linewidth]{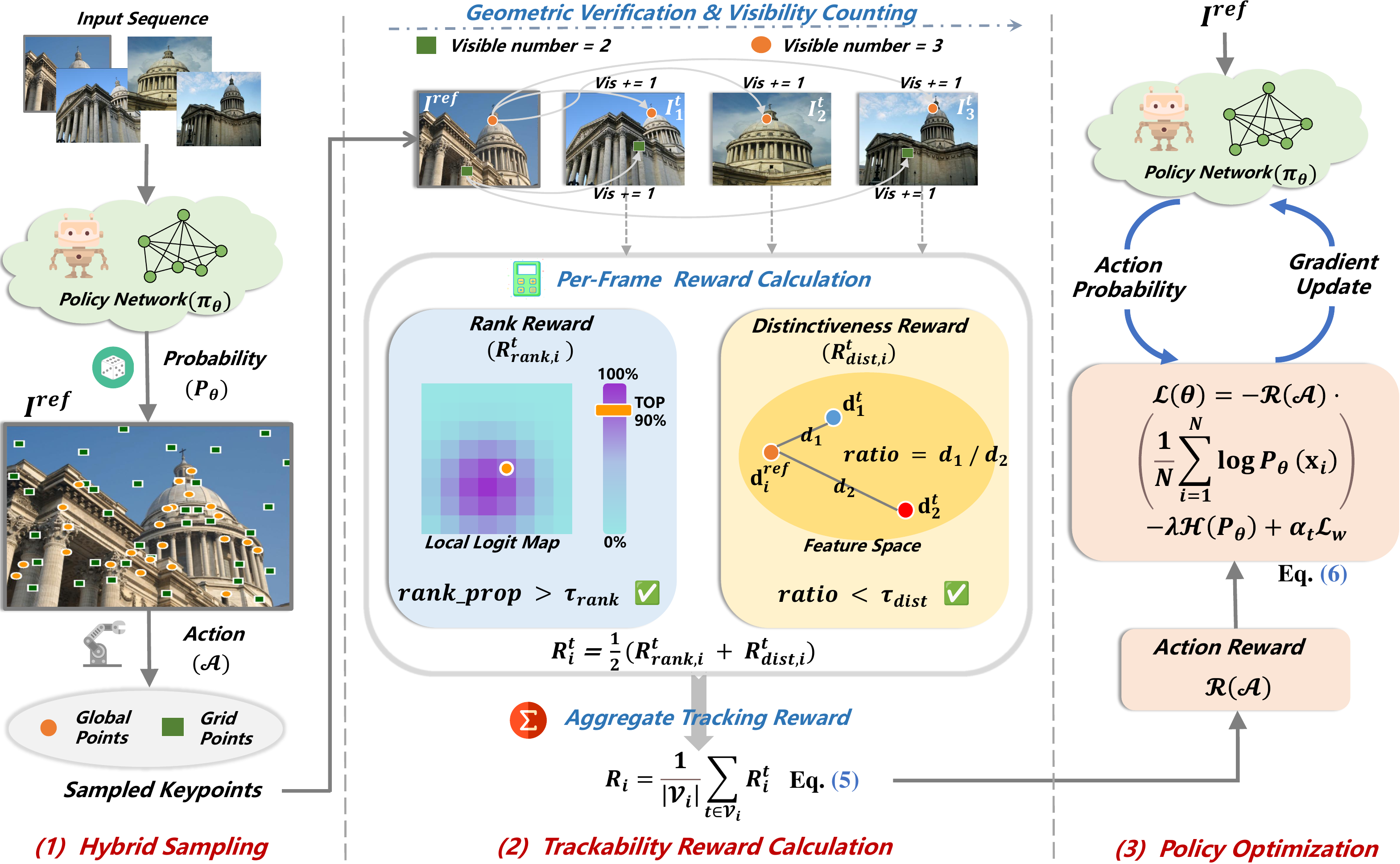}
    \caption{ Overview of our proposed Sequence-Aware Keypoint Policy Learning framework: First, we select a reference frame from the input image sequence and perform hybrid keypoint sampling on it. Next, we leverage geometric mapping to locate the corresponding positions of the reference frame’s keypoints in other frames of the sequence. We then count the number of these keypoints visible across the entire sequence. After that, we compute per-frame rewards for each sampled point and aggregate these into a final track reward. Finally, we update the policy network’s gradients.}
    \label{fig:main_frame}
\vspace{-0.9 em}
\end{figure*}

Our TraqPoint utilizes a lightweight policy network and a  sequence-aware reward to select keypoints explicitly optimized for long-term trackability. In Section~\ref{sec:Preliminaries}, we first introduce the  preliminaries, detailing the dual-branch network architecture (Fig.~\ref{fig:network}) and the pre-training of the descriptor branch, which provides a stable signal for reward calculation. Subsequently, we formally define the Problem Formulation (Sec.~\ref{sec:Problem}) of our RL approach. We then detail our core innovations: hybrid sampling strategy (Sec.~\ref{sec:Sampling}) and the trackability reward (Sec.~\ref{sec:Reward}). Finally, we describe the policy optimization process (Sec.~\ref{sec:Optimization}) for the network. Fig.~\ref{fig:main_frame} illustrates the complete RL training process.

\subsection{Preliminaries}
\label{sec:Preliminaries}

Our method builds upon the recent success of dual-branch architectures~\cite{edstedt2024dedodev2, chen2025rdd}. We adopt a ``describe-then-detect" training paradigm, where the descriptor branch is pre-trained first and then kept frozen in keypoint branch training. This two-stage strategy allows our framework to focus exclusively on optimizing the keypoint detection policy, which is the core of our work.

\smallskip\noindent\textbf{Network Architecture.} As shown in Fig.~\ref{fig:network}, we adopt the same network architecture as RDD~\cite{chen2025rdd}. The descriptor branch consists of a backbone and a multi-scale deformable transformer module. We replace the original ResNet-50 backbone with a DINOv3-ConvNeXt (base)~\cite{simeoni2025dinov3} and utilize its features from four different scales. The transformer module then aggregates these features to produce a dense descriptor map $D \in \mathbb{R}^{\frac{H}{4} \times \frac{W}{4} \times 256}$. The keypoint branch retains the lightweight 4-layer convolutional design from ALIKED~\cite{zhao2023aliked}. As this branch serves as the policy network in our RL framework, we term its final $1 \times 1$ convolutional layer the policy head, which outputs the final logit map $L \in \mathbb{R}^{H \times W}$.

While this backbone upgrade improves performance (see ablation in Sec.~\ref{sec:ablation}), our primary contribution is the novel RL training paradigm, not the architectural modifications.

\smallskip\noindent\textbf{Descriptor Branch Pre-training.}
Following prior work~\cite{potje2024xfeat,chen2025rdd}, we train the descriptor branch $\Phi$ on image pairs from the MegaDepth dataset~\cite{li2018megadepth}. We first leverage camera poses and depth maps to generate ground-truth pixel correspondences $\mathcal{M}_{\text{gt}}$. For a given pair $(I_A, I_B)$, we sample the corresponding sets of descriptors $\mathbf{d}_A$ and $\mathbf{d}_B$ from the outputs of $\Phi$.
We then compute the similarity matrix $\mathbf{S}$ between $\mathbf{d}_A$ and $\mathbf{d}_B$, and apply a dual-softmax operation to obtain the probability matrix $\mathbf{P}$. To optimize the network, we apply the same focal loss function ~\cite{chen2025rdd} on the positive correspondences (diagonal elements $\mathbf{P}_{ii}$) :
\begin{equation}
\mathcal{L}_{desc} = -\frac{1}{|\mathcal{M}_{\text{gt}}|} \sum_{i \in \mathcal{M}_{\text{gt}}} \alpha (1 - \mathbf{P}_{ii})^\gamma \log(\mathbf{P}_{ii}),
\end{equation}
where $\alpha=0.25$ and $\gamma=2$. Upon completion of this training stage, we freeze the parameters of $\Phi$. 

\subsection{Problem Setup}
\label{sec:Problem}

To address the dual challenges of the discrete selection problem inherent in keypoint detection and the objective misalignment of pairwise training, we introduce a reinforcement learning framework that reformulates keypoint detection as a sequential decision-making task. 
In this framework, the policy network $\pi_\theta$ (parameterized by $\theta$) acts as the agent. The state $s$ is defined as the reference image $I^{ref}$. Given $s$, 
the policy outputs a pixel-wise probability distribution, denoted as
$P_\theta = \pi_\theta(s)$, over the image spatial domain.
An action is defined as sampling a set of $N$ keypoints independently from this distribution: $\mathcal{A} = \{ \mathbf{x}_i \}_{i=1}^N$, where $\mathbf{x}_i \sim P_\theta $.
The objective is to maximize the expected  reward $\mathcal{R}(\mathcal{A})$, which is the average long-term trackability score $R_i$ for each keypoint in $\mathcal{A}$ (detailed in Sec.~\ref{sec:Reward}). This is formulated as maximizing the objective function:
\begin{equation}
\mathcal{J}(\theta) = \mathbb{E}_{\mathcal{A} \sim P_\theta} [\mathcal{R}(\mathcal{A})].
\end{equation}

\subsection{Hybrid Sampling Strategy}
\label{sec:Sampling}
While the action set $\mathcal{A}$ is sampled from the policy's output distribution $P_\theta$, the specific sampling method significantly impacts performance. A naive approach of sampling from the global distribution can cause keypoints to cluster in high-probability regions, resulting in poor spatial coverage. We therefore propose a hybrid sampling strategy to balance exploitation and exploration (via spatial coverage). Specifically, $\mathcal{A}$ is a union of two subsets:
\begin{itemize}
    \item \textbf{Global Sampling:} We sample $N_g$ points by drawing directly from the global distribution $P_\theta$, focusing on high-probability regions.
    \item \textbf{Grid Sampling:} We divide the image into a $G \times G$ regular grid (resulting in $N_{\text{grid}} = G^2$). For each cell, we first extract the local logit submap from $L$ and apply spatial softmax to obtain a local policy distribution. We then sample one keypoint from each cell based on this local distribution, ensuring spatial coverage while selecting locally optimal points.
\end{itemize}

This hybrid approach generates a spatially diverse set of $N = N_g + N_{\text{grid}}$ keypoints. For the policy gradient calculation, the probability of {every} sampled keypoint $\mathbf{x}_i \in \mathcal{A}$ (regardless of its source) is defined by its value in the {global distribution $P_\theta(\mathbf{x}_i)$}.

\subsection{Trackability Reward Formulation}
\label{sec:Reward}
The core of our framework is the reward function $R_i$ for each sampled keypoint $\mathbf{x}_i \in \mathcal{A}$ (located in $I^{ref}$). For each keypoint $\mathbf{x}_i$, we project it into every target frame $I^t$ using the known relative pose and depth, yielding a projected coordinate $\mathbf{x}_i^t$. A projection is considered {visible} if it falls within the image bounds and passes standard geometric verification checks (such as depth consistency). We denote the set of frame indices $t$ where $\mathbf{x}_i$ is visible as $\mathcal{V}_i$.

For each visible projection $\mathbf{x}_i^t$ (where $t \in \mathcal{V}_i$), we compute a per-frame reward $R_i^t$. To prevent sparse signals, both components of this reward are designed as {linear scores} rather than binary pass/fail signals. This $R_i^t$ combines two complementary signals:

\smallskip\noindent\textbf{Rank Reward ($R_{\text{rank}}$).} 
This reward quantifies the cross-view consistency of saliency for high-probability points selected from the $I^{ref}$. 
In the target frame $I^t$, we compute the logit map $L^t = \pi_\theta(I^t)$. We assess the logit value $L^t(\mathbf{x}_i^t)$ relative to its local $K \times K$ patch. We define $\text{rank\_prop}$ as the {saliency percentile score}, representing the proportion of points it is more salient than. For example, in a $10 \times 10$ patch (100 points), the 10th-highest logit value has a $\text{rank\_prop}$ of 0.9 (as it is more salient than 90\% of the points). We set a cutoff threshold $\tau_{\text{rank}}=0.2$, meaning only points that are more salient than the bottom 20\% of points  receive a reward. This reward is defined as follows:
\begin{equation}
    R_{\text{rank}, i}^t = \max\left(0, \frac{\text{rank\_prop} - \tau_{\text{rank}}}{1.0 - \tau_{\text{rank}}}\right).
\end{equation}

\smallskip\noindent\textbf{Distinctiveness Reward ($R_{\text{dist}}$).} This reward, inspired by Lowe's ratio test~\cite{lowe2004distinctive}, measures the distinctiveness of the selected point. We use the frozen descriptor branch $\Phi$ to get the reference descriptor $\mathbf{d}_i = \Phi(I^{ref}, \mathbf{x}_i)$. We then compare $\mathbf{d}_i$ against the set of {all $N$} projected descriptors in the target frame, $\{\mathbf{d}_j^t = \Phi(I^t, \mathbf{x}_j^t) \mid j=1...N \text{ and } t \in \mathcal{V}_j \}$. Let $d_1$ and $d_2$ be the distances to the nearest and second-nearest neighbors of $\mathbf{d}_i$ {within this set of $N$ descriptors}. We define $\text{ratio} = d_1 / d_2$. We reward points only if this $\text{ratio}$ is below our threshold $\tau_{\text{dist}}=0.85$. This reward is  linearly scaled:
\begin{equation}
    R_{\text{dist}, i}^t = \max\left(0, \frac{\tau_{\text{dist}} - \text{ratio}}{\tau_{\text{dist}}}\right).
\end{equation}

The per-frame reward $R_i^t$ is the average of these two components. The final trackability reward for keypoint $\mathbf{x}_i$ is $R_i$:
\begin{equation}
    R_i = \frac{1}{|\mathcal{V}_i|} \sum_{t \in \mathcal{V}_i} R_i^t.
\end{equation}
If a keypoint is never visible ($|\mathcal{V}_i| = 0$), its reward $R_i$ is zero.

\begin{figure*}[ht]
\vspace{-0.8 em}
    \centering
\includegraphics[width=0.98\linewidth]{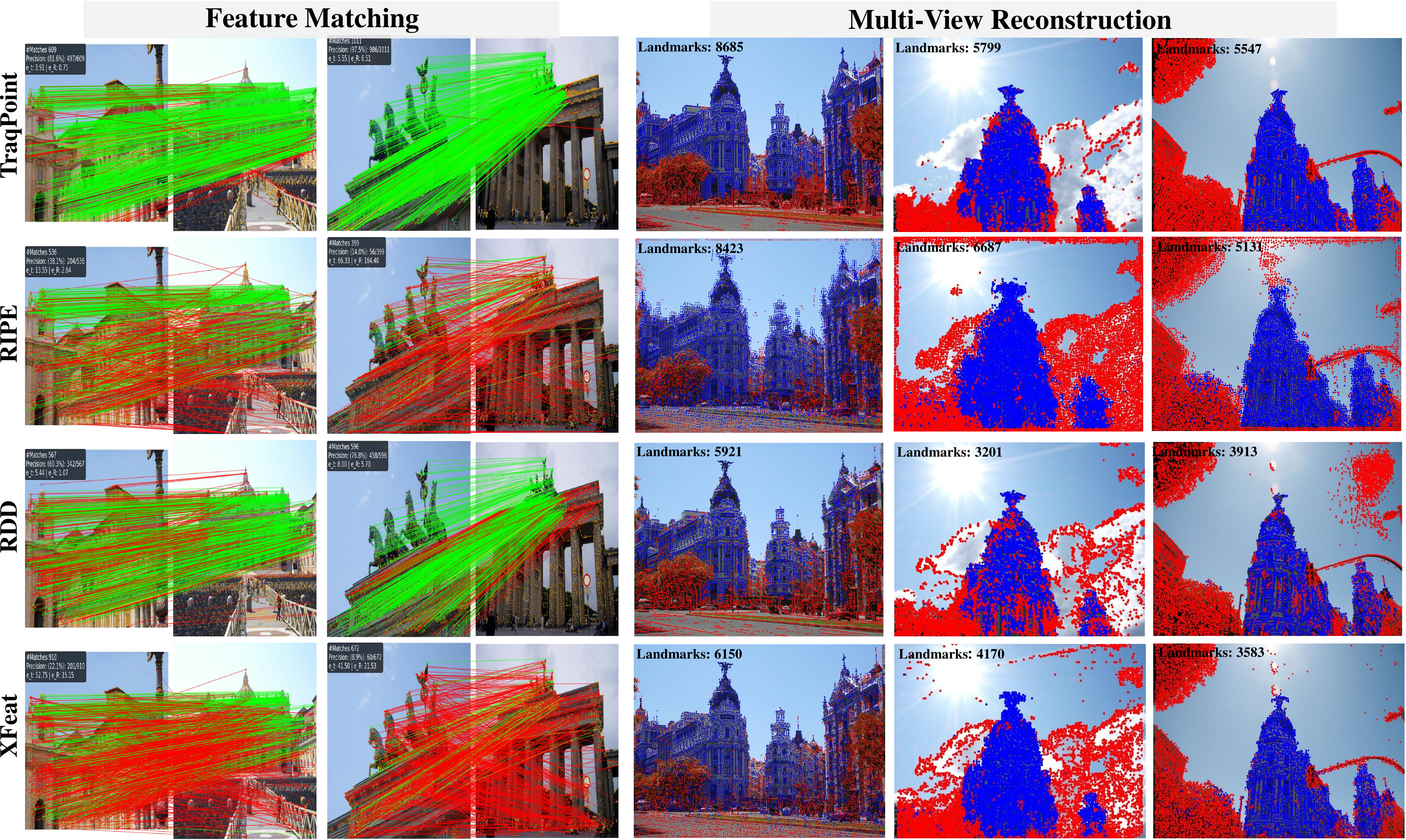}
\vspace{-0.5 em}
    \caption{Qualitative results on the MegaDepth dataset ~\cite{li2018megadepth} and the ETH benchmark~\cite{schonberger2017comparative}. For feature matching, keypoints are plotted in \textcolor{orange}{orange}. {Green lines} indicate correct matches, while {red lines} denote incorrect ones. For multi-view reconstruction, keypoints that successfully generate point cloud landmarks are marked in \textcolor{blue}{blue}; failed ones are marked in \textcolor{red}{red}. Our TraqPoint produces keypoints with better structural saliency and high cross-view consistency, which effectively improves feature matching and reconstruction tasks.}
\label{fig:visulize}
\vspace{-0.9 em}
\end{figure*}

\subsection{Policy Optimization}
\label{sec:Optimization}

We optimize the policy network $\pi_\theta$ using a composite loss function. The primary component is the policy gradient term. 
To reduce variance, we use the average reward of the entire action set.
Let $\mathcal{R}(\mathcal{A}) = \frac{1}{N} \sum_{i=1}^{N} R_i$. 
The policy gradient is then estimated using the average log-probability of all sampled actions, $\log P_\theta(\mathbf{x}_i)$ (from Sec.~\ref{sec:Sampling}). Second, to encourage spatial diversity and prevent mode collapse, we add a spatial entropy regularization term, $\mathcal{H}(P_\theta) = -\sum_{\mathbf{x}} P_\theta(\mathbf{x}) \log P_\theta(\mathbf{x})$. Finally, to accelerate convergence in the early stages, we employ a warm-up loss $\mathcal{L}_{w}$ for the initial 10\% of training epochs. This term uses keypoint locations from the FAST detector~\cite{rosten2008faster} to weakly supervise the $\text{sigmoid}$ of the policy's logit map (\eg, via a BCE loss). The final loss function is:
\begin{equation}
\label{eq:policy_loss}
    \mathcal{L}(\theta) = - \mathcal{R}(\mathcal{A}) \cdot \left( \frac{1}{N} \sum_{i=1}^{N} \log P_\theta(\mathbf{x}_i) \right) - \lambda \mathcal{H}(P_\theta) +  \alpha_t \mathcal{L}_{w},
\end{equation}
where $\lambda$ is a fixed hyperparameter that balances the entropy term with the main loss, and is set to 0.001.
The warm-up weight $\alpha_t$ linearly anneals from 1.0 to 0.0 during the first 10\% of epochs, after which only the RL loss remains.

\section{Experiments}
\vspace{-0.1 em}
\label{sec:experiments}
To fully validate the effectiveness of our proposed TraqPoint, we conduct evaluations on pair-level tasks (relative pose estimation, visual localization), and sequence-level tasks (visual odometry, 3D reconstruction). Furthermore, we perform ablation experiments to dissect the contributions of the key modules in our framework.
\subsection{Training Data}
\vspace{-0.1 em}
We construct sequential training data from the MegaDepth dataset~\cite{li2018megadepth} for  keypoint branch training. Specifically, we first randomly select a reference image from each sub-scene, then identify the remaining images that have an overlap ratio  between 10\% and 70\% with the reference via their depth maps and camera poses, and finally randomly sample images from this overlapping subset to form a training sequence with the reference. Additional details about the data construction are provided in the supplementary material.
\vspace{-0.3 em}
\subsection{Implementation Details}
\label{sec:details}
\vspace{-0.1 em}
 For descriptor pre-training, images are resized to 800px. For keypoint policy learning, images are resized to 480px, and we use fixed-length sequences of 5 images with $N=256$ keypoints sampled per RL step. 
 Both stages are trained for 50,000 steps on 8 NVIDIA H20 GPUs.
 We use Adam optimizer with a cosine annealing learning rate scheduled from $2 \times 10^{-4}$ down to $5 \times 10^{-6}$. 
 During inference, we apply a sigmoid function to the output of keypoint branch and use Non-Maximum Suppression (NMS) with a $5 \times 5$ window to extract keypoints.
Following ~\cite{tyszkiewicz2020disk, chen2025rdd}, we utilize the soft mutual nearest neighbor (MNN) operation, which computes matching probabilities by applying softmax over both dimensions of the descriptor similarity matrix. We discard matches with probability  below 0.01 for feature matching.
A comparison of the runtime between our method and other approaches is provided in the supplementary material.

\subsection{Relative Pose Estimation}
\label{sub:Pose}
\vspace{-0.4 em}
\smallskip\noindent\textbf{Datasets.} 
Following prior works~\cite{potje2024xfeat, chen2025rdd}, we use the MegaDepth~\cite{li2018megadepth} and ScanNet~\cite{dai2017scannet} test sets to evaluate the relative pose estimation task. These datasets feature outdoor and indoor scenes respectively, and their main challenges include large viewpoint changes, illumination variations, and repetitive textures. There is no overlap between these test sets and our training set.

\smallskip\noindent\textbf{Metrics and Compared Methods.}
We report the Area Under the Curve (AUC) for pose errors at thresholds of 5°, 10°, and 20°. To ensure reliable pose estimation, we use the RANSAC function of OpenCV~\cite{bradski2000opencv} to solve for the essential matrix.
We compared TraqPoint against the state-of-the-art detector/descriptor
methods~\cite{detone2018superpoint,tyszkiewicz2020disk,zhao2023aliked,edstedt2024dedodev2,potje2024xfeat,chen2025rdd,kunzel2025ripe}, and learned matching methods~\cite{sarlin2020superglue,lindenberger2023lightglue}.
For all methods, images are resized to a maximum dimension of 1,600 pixels for the MegaDepth dataset, while the VGA resolution is used for the ScanNet dataset. All methods sample the top-4096 keypoints, and conduct experiments following the optimal settings from their official repositories.

\begin{table}[!t]
\small
 \centering
 \caption{\textbf{Comparison on the MegaDepth~\cite{li2018megadepth} and ScanNet~\cite{dai2017scannet}}. Top 4,096 features are used for all sparse matching methods. Best in bold, second best underlined.}
 \resizebox{0.99\columnwidth}{!}{
 \begin{tabular}{lcccccc}
    \hline      
    \multirow{3}{*}{\textbf{Method}} &\multicolumn{3}{c}{\textbf{MegaDepth-1500}} & \multicolumn{3}{c}{\textbf{ScanNet}}  \\ 
                                     & \multicolumn{3}{c}{AUC $\uparrow$}    &   \multicolumn{3}{c}{AUC $\uparrow$}    \\
                                     & @5\degree & @10\degree & @20\degree    &      @5\degree & @10\degree & @20\degree      \\
    
    \midrule\hline
    \textbf{\emph{Sparse with Learned Matcher}} & & & & & & \\
    \rowcolor{mysectioncolor} SP~\cite{detone2018superpoint}+SG~\cite{sarlin2020superglue}~\tiny{CVPR'19} & 49.7 & {67.1} & {80.6} & {16.1} & {33.4} & {50.2} \\      
    SP~\cite{detone2018superpoint}+LG~\cite{lindenberger2023lightglue}~\tiny{ICCV'23} & {49.9} & 67.0 & 80.1 & 16.2 & 33.0 & 49.7 \\
    \midrule
    \textbf{\emph{Sparse with MNN}} & & & & & & \\
    \rowcolor{mysectioncolor} SuperPoint~\cite{detone2018superpoint}~\tiny{CVPRW'18} & 24.1 & 40.0 & 54.7 & 7.0 & 17.2 & 29.8 \\
    DISK~\cite{tyszkiewicz2020disk}~\tiny{NeurIps'20} & 38.5 & 53.7 & 66.6 & 8.2 & 18.3 & 30.7 \\
    \rowcolor{mysectioncolor} ALIKED~\cite{zhao2023aliked}~\tiny{TIM'23} & 41.8 & 56.8 & 69.6 & 9.7 & 22.0 & 37.1 \\
    XFeat~\cite{potje2024xfeat}~\tiny{CVPR'24} & 29.6 & 45.8 & 60.4 & 6.9 & 16.7 & 30.4 \\
    \rowcolor{mysectioncolor} DeDoDe-V2-G~\cite{edstedt2024dedodev2}~\tiny{CVPRW'24} & {49.7} & {66.6} & {79.4} & {13.2} & {27.7} & {43.7}\\
    RDD~\cite{chen2025rdd}~\tiny{CVPR'25} & \underline{51.9} & \underline{68.0} & \underline{79.9} & \underline{13.7} & \underline{29.3} & \underline{45.3} \\
    \rowcolor{mysectioncolor} RIPE~\cite{kunzel2025ripe}~\tiny{ICCV'25} & {45.4} & {60.8} & {72.9} & {9.4} & {20.5} & {33.5} \\
    \rowcolor{myourscolor} TraqPoint (Ours) & \textbf{55.8} & \textbf{71.3} & \textbf{83.0} & \textbf{16.6} & \textbf{32.8} & \textbf{49.5} \\
 \bottomrule
 \end{tabular}
 }
 \label{tab:pose_estimate}
\vspace{-1.0 em}
\end{table}

\smallskip\noindent\textbf{Results.}
Tab.~\ref{tab:pose_estimate} shows our TraqPoint outperforms some state-of-the-art pairwise and RL-based methods across the MegaDepth and ScanNet benchmarks. 
Specifically, compared to RDD~\cite{chen2025rdd}, we improve the AUC@5 by 3.9 and 2.9 on MegaDepth and ScanNet, respectively.
We also gain significant matching accuracy advantages over RL-based methods including DISK~\cite{tyszkiewicz2020disk} and RIPE~\cite{kunzel2025ripe}. Even with only the MNN matcher, our method delivers a 5.9 boost in AUC@5 boost on MegaDepth compared to methods equipped with extra matchers (\eg, SP+LG~\cite{lindenberger2023lightglue}), and achieves comparable accuracy on ScanNet—notably, our model is never trained on indoor data. 
As shown in Fig.~\ref{fig:visulize}, our method generates more structurally salient keypoints in texture-rich regions with strong consistency even under extreme viewpoint variations, which further boosts the accuracy of pairwise matching.

\vspace{-0.2 em}
\subsection{Visual Localization}
\label{sub:visual local}
\vspace{-0.4 em}
\begin{table}[!t]
    \centering
    \caption{{\bf Visual Localization on Aachen day-night~\cite{sattler2018benchmarking}.} Best in bold, second best underlined.}
    \resizebox{0.9\columnwidth}{!}{%
    \begin{tabular}{lcccc}
        \toprule
        \multirow{2}{1.3cm}[-.4em]{Methods}
        & Day & & Night\\
        \cmidrule(lr){2-4}
        & \multicolumn{3}{c}{(0.25m,2°) / (0.5m,5°) / (1.0m,10°)}\\
        \midrule\hline
        \rowcolor{mysectioncolor} SuperPoint~\cite{detone2018superpoint}~\tiny{CVPRW'18}     & \underline{87.4} / 93.2 / 97.0 && 77.6 / 85.7 / 95.9 \\
        DISK~\cite{tyszkiewicz2020disk}~\tiny{NeurIps'20}            & 86.9 / \textbf{95.1} / \underline{97.8} && 83.7 / 89.8 / \underline{99.0} \\
        \rowcolor{mysectioncolor} ALIKE~\cite{Zhao2022ALIKE}~\tiny{TM'22}            & 85.7 / 92.4 / 96.7 && 81.6 / 88.8 / \underline{99.0} \\
        ALIKED~\cite{zhao2023aliked}~\tiny{TIM'23}           & 86.5 / 93.4 / 96.8 && \underline{85.7} / \underline{91.8} / 96.9 \\
        \rowcolor{mysectioncolor} XFeat~\cite{potje2024xfeat}~\tiny{CVPR'24}           & 84.7  / 91.5 / 96.5 && 77.6 / 89.8 / {98.0} \\
        RDD~\cite{chen2025rdd}~\tiny{CVPR'25} & {87.0}  / \underline{94.2} / \underline{97.8} && \textbf{86.7} / \textbf{92.9} /
        \underline{99.0} \\
        \rowcolor{myourscolor} TraqPoint (Ours) & \textbf{87.9}  / \textbf{95.1} / \textbf{97.9} && \underline{85.7} / \textbf{92.9} /
        \textbf{100.0} \\
        
        \bottomrule
    \end{tabular}
    }
    \label{tab:aachen}
\vspace{-0.9 em}
\end{table}

\smallskip\noindent \textbf{Dataset.}
We validate TraqPoint’s performance on visual localization using the Aachen Day-Night dataset~\cite{sattler2018benchmarking}. This dataset is specifically designed to localize high-quality nighttime images against a daytime 3D model, and it comprises 14,607 images covering variations in weather, season, viewpoint, and day-night cycles.

\smallskip\noindent \textbf{Metrics and Compared Methods.}
Following~\cite{potje2024xfeat,chen2025rdd}, we use HLoc~\cite{sarlin2019coarse} to evaluate all methods~\cite{potje2024xfeat, zhao2023aliked, detone2018superpoint, tyszkiewicz2020disk, chen2025rdd}. Images are resized to a maximum side length of 1,024 pixels, and we sample the top 4,096 keypoints for all approaches. Metrics are computed using the evaluation tool from~\cite{sattler2018benchmarking}, including the AUC of estimated camera poses under threshold of $\{0.25$m$, 0.5$m$, 1.0$m$\}$ for translation errors and $\{2\degree, 5\degree, 10\degree\}$ for rotation errors respectively.  

\smallskip\noindent \textbf{Results.}
Tab.~\ref{tab:aachen} presents the results of visual localization. TraqPoint achieves the best performance across all daytime settings and the two nighttime settings, further demonstrating its robustness.

\vspace{-0.2 em}
\subsection{Visual Odometry (VO)}
\label{sub:vo}
\vspace{-0.4 em}
\begin{table}[!t]
    \centering
    \caption{ \textbf{Visual Odometry test results on sequence (01-03) of KITTI~\cite{geiger2012we}}. ATE: Average Trajectory Error; MTE: Maximum Trajectory Error; AKTL: Average Keypoint Tracking Length. }
    \resizebox{0.7\columnwidth}{!}{
    \begin{tabular}{l c c c  }
        \midrule
        \rowcolor{myheadercolor}
        Methods & ATE $\downarrow$ & MTE $\downarrow$ & AKTL $\uparrow$   \\
        \hline\hline
        \multicolumn{4}{l}{\textbf{Seq-01}} \\
        \rowcolor{mysectioncolor} RootSIFT~\cite{arandjelovic2012three}~\tiny{CVPR'12}       & 60.4  & 102.1 & 4.4   \\
        XFeat~\cite{potje2024xfeat}~\tiny{CVPR'24}   &  55.0 & 92.7 & 3.9   \\
        \rowcolor{mysectioncolor} RDD~\cite{chen2025rdd}~\tiny{CVPR'25}   & \underline{35.3}  & \underline{78.3} & \underline{4.6}   \\
        RIPE~\cite{kunzel2025ripe}~\tiny{ICCV'25}   & 43.7  & 90.6 & 4.1   \\
        \rowcolor{myourscolor} TraqPoint (Ours) & \textbf{29.9} & \textbf{51.4} & \textbf{7.3}   \\
        \hline
        \multicolumn{4}{l}{\textbf{Seq-02}} \\
        \rowcolor{mysectioncolor} RootSIFT~\cite{arandjelovic2012three}~\tiny{CVPR'12}       & 24.6 & 45.9 & 2.0   \\
        XFeat~\cite{potje2024xfeat}~\tiny{CVPR'24}   &  36.5  & 87.4  &  1.8  \\
        \rowcolor{mysectioncolor} RDD~\cite{chen2025rdd}~\tiny{CVPR'25}   & \underline{12.8}  & \underline{40.3} & \underline{2.4}   \\
        RIPE~\cite{kunzel2025ripe}~\tiny{ICCV'25}   & 35.4 & 82.3 & 1.7   \\
        \rowcolor{myourscolor} TraqPoint (Ours) & \textbf{11.8} & \textbf{22.1} & \textbf{3.8}   \\
        \hline
        \multicolumn{4}{l}{\textbf{Seq-03}} \\
        \rowcolor{mysectioncolor} RootSIFT~\cite{arandjelovic2012three}~\tiny{CVPR'12}       & 5.8  & 16.6 & {4.8}    \\
        XFeat~\cite{potje2024xfeat}~\tiny{CVPR'24}   &  11.7 & 26.8 & 4.2   \\
        \rowcolor{mysectioncolor} RDD~\cite{chen2025rdd}~\tiny{CVPR'25}   & {4.0}  & {11.2} & \underline{5.2}   \\
        RIPE~\cite{kunzel2025ripe}~\tiny{ICCV'25}   & \underline{3.9}  & \underline{9.4} & 4.8   \\
        \rowcolor{myourscolor} TraqPoint (Ours) & \textbf{1.3} & \textbf{2.9} & \textbf{8.7}   \\
        \hline
    \end{tabular}
    }
    \label{tab:kitti_vo}
\vspace{-0.9 em}
\end{table}

\smallskip\noindent \textbf{Dataset.}
Visual Odometry (VO) is a core component of SLAM systems, serving as a direct testbed for long-term keypoint trackability. 
We thus evaluate TraqPoint on the KITTI Odometry Benchmark~\cite{geiger2012we}, which provides 22 stereo image sequences covering diverse real-world driving scenarios (\eg, varying illumination, dynamic objects). Among these sequences, 10 are accompanied by ground-truth trajectories annotated via high-precision GPS/IMU for quantitative evaluation. 
We report the results of sequences 01–03, while the results of the remaining sequences along with detailed definitions of the metrics are included in the supplementary material.

\smallskip\noindent \textbf{Metrics and Compared Methods.}
We evaluate odometry benchmark from two perspectives: trajectory estimation accuracy and keypoint tracking consistency. For trajectory accuracy, we compute Average Trajectory Error \textbf{(ATE)} and Maximum Trajectory Error \textbf{(MTE)} by calculating 2D positional errors of projected $x$-$z$ plane trajectories. 
For keypoint tracking consistency, Average Keypoint Tracking Length \textbf{(AKTL)} is defined as the mean tracking length of valid keypoints.
We compared TraqPoint against the handcrafted features~\cite{arandjelovic2012three} and  learnable detectors~\cite{potje2024xfeat,chen2025rdd,kunzel2025ripe}. 
All methods use the top-4096 keypoints for prediction, and adopt mutual nearest neighbor (MNN) matching.

\smallskip\noindent \textbf{Results.}
Tab.~\ref{tab:kitti_vo} presents our performance on the odometry task. Our method achieves the best results in both Average Trajectory Error and Maximum Trajectory Error. Notably, it delivers remarkable gains in the Average Keypoint Trajectory Length (AKTL) metric—outperforming RDD~\cite{chen2025rdd} and RIPE~\cite{kunzel2025ripe} with significant margins. 
Even under outdoor fast-motion and associated challenges (rapid texture changes,  dynamic objects), our keypoints still maintain good spatial consistency in temporal sequences, enabling more stable camera pose and trajectory estimation.

\begin{table}[!t]
    \centering
    \caption{\textbf{Results on ETH benchmark~\cite{schonberger2017comparative} for 3D reconstruction.} Best in bold, second best underlined.}
    \resizebox{0.96\columnwidth}{!}{
    \renewcommand\arraystretch{1.2}
    \begin{tabular}{l c c c c }
        \toprule
        \rowcolor{myheadercolor} 
        Methods & Reg. Img$\uparrow$ & Sparse Pts$\uparrow$ & Track Len$\uparrow$ & Reproj Err$\downarrow$  \\
        \midrule\hline
        \multicolumn{5}{l}{\textbf{Madrid Metropolis (1344 imgs)}} \\
        \rowcolor{mysectioncolor} RootSIFT~\cite{arandjelovic2012three}~\tiny{CVPR'12}       & 500  & 116k & 6.32 & \textbf{0.60px}  \\
        SuperPoint~\cite{detone2018superpoint}~\tiny{CVPRW'18}     & 438  & 29k  & 9.03 & 1.05px  \\
        \rowcolor{mysectioncolor} D2-Net~\cite{dusmanu2019d2}~\tiny{CVPR'19}    & 501  & 84k & 6.33 & 1.28px  \\
        ASLFeat~\cite{luo2020aslfeat}~\tiny{CVPR'20}        & 613  & 96k  & 8.76 & 0.90px  \\
        \rowcolor{mysectioncolor} PosFeat~\cite{li2022decoupling}~\tiny{CVPR'22}        & 419  & 72k & 9.18 & \underline{0.86px}  \\
        XFeat~\cite{potje2024xfeat}~\tiny{CVPR'24}        & 581  & 111k & 6.16 &  1.27px \\
        \rowcolor{mysectioncolor} RDD~\cite{chen2025rdd}~\tiny{CVPR'25}        & 632  & 154k & \underline{9.40} &  1.12px \\
        RIPE~\cite{kunzel2025ripe}~\tiny{ICCV'25}        & \underline{644}  & \underline{167k} & 5.87 &  1.25px\\
        \rowcolor{myourscolor} TraqPoint (Ours)& \textbf{693} & \textbf{254k} & \textbf{11.14} & 1.03px  \\
        \hline
        \multicolumn{5}{l}{\textbf{Gendarmen-markt (1463 imgs)}} \\
        \rowcolor{mysectioncolor} RootSIFT~\cite{arandjelovic2012three}~\tiny{CVPR'12}       & 1035 & \underline{338k} & 5.52 & \textbf{0.69px}  \\
        SuperPoint~\cite{detone2018superpoint}~\tiny{CVPRW'18}     & 967  & 93k  & 7.22 & 1.03px  \\
        \rowcolor{mysectioncolor} D2-Net~\cite{dusmanu2019d2}~\tiny{CVPR'19}     & 1053  & 250k & 5.08 & 1.19px  \\
        ASLFeat~\cite{luo2020aslfeat}~\tiny{CVPR'20}        & 1040 & 221k & 8.72 & 1.00px  \\
        \rowcolor{mysectioncolor} PosFeat~\cite{li2022decoupling}~\tiny{CVPR'22}        & 956  & 240k & 8.40 & \underline{0.92px}  \\
        XFeat~\cite{potje2024xfeat}~\tiny{CVPR'24}        &  1049 & 276k & 6.74 &  1.30px \\
        \rowcolor{mysectioncolor} RDD~\cite{chen2025rdd}~\tiny{CVPR'25}        &  1062 & 309k & \underline{9.87} &  1.17px \\
        RIPE~\cite{kunzel2025ripe}~\tiny{ICCV'25}        &  \underline{1069} & 321k & 7.21 & 1.26px \\
        \rowcolor{myourscolor} TraqPoint (Ours)& \textbf{1093} & \textbf{401k} & \textbf{11.06} & 1.08px  \\
        \hline
        \multicolumn{5}{l}{\textbf{Tower of London (1576 imgs)}} \\
        \rowcolor{mysectioncolor} RootSIFT~\cite{arandjelovic2012three}~\tiny{CVPR'12}       & 804  & 239k & 7.76 & \textbf{0.61px}  \\
        SuperPoint~\cite{detone2018superpoint}~\tiny{CVPRW'18}     & 681  & 52k  & 8.67 & 0.96px  \\
        \rowcolor{mysectioncolor} D2-Net~\cite{dusmanu2019d2}~\tiny{CVPR'19}   & 785  & 180k & 5.32 & 1.24px  \\
        ASLFeat~\cite{luo2020aslfeat}~\tiny{CVPR'20}        & 821  & 222k & {12.16} & 0.92px  \\
        \rowcolor{mysectioncolor} PosFeat~\cite{li2022decoupling}~\tiny{CVPR'22}        & 778  & {262k} & 11.64 & \underline{0.90px}  \\
        XFeat~\cite{potje2024xfeat}~\tiny{CVPR'24}        & 796  & 244k & 8.50 & 1.28px  \\
        \rowcolor{mysectioncolor} RDD~\cite{chen2025rdd}~\tiny{CVPR'25}        & \underline{834}  & \underline{278k} & \underline{12.20} &  1.15px \\
        RIPE~\cite{kunzel2025ripe}~\tiny{ICCV'25}        &  823 & 269k & 9.14 & 1.25px \\
        \rowcolor{myourscolor} TraqPoint (Ours) & \textbf{875} & \textbf{341k} & \textbf{13.28} & 1.06px  \\
        \hline
    \end{tabular}
    }
    \label{tab:eth_3d_recon_simplified}
\vspace{-0.9 em}
\end{table}

\vspace{-0.2 em}
\subsection{3D Reconstruction}
\label{sub:rec}
\vspace{-0.4 em}
\smallskip\noindent \textbf{Dataset.}
We use the ETH benchmark \cite{schonberger2017comparative} to demonstrate effectiveness on 3D reconstruction tasks. Following prior works \cite{luo2020aslfeat, li2022decoupling}, we evaluate on three medium-scale scenes from this benchmark.

\smallskip\noindent \textbf{Metrics and Compared Methods.}
For each scene, we perform matching with a ratio test of 0.8 and mutual check for outlier rejection, followed by SfM algorithm execution using COLMAP~\cite{schonberger2016structure}.
We then report the number of registered images (Reg. Images), the number of sparse points (Sparse Pts), average track length (Track Len) and mean reprojection error (Reproj. Error).  The maximum number of keypoints is limited to 20k for all methods.

\smallskip\noindent \textbf{Results.}
Tab.~\ref{tab:eth_3d_recon_simplified} demonstrates the effectiveness of our method in multi-view sparse reconstruction. Our approach achieves the best performance across key metrics: the number of registered images, generated point cloud number, and keypoint track length. 
The marginal increase in reprojection error stems from retaining ``hard'' keypoints (\eg, under large viewpoint changes), which are essential for the significant gains in reconstruction density and track length.
These results reflect that the strong consistency of our keypoints effectively boosts reconstruction quality. 
Further visualizations in Fig.~\ref{fig:visulize} demonstrate that our keypoints yield a more rational distribution: they are less likely to fall into texture-less regions (\eg, sky) and instead exhibit a more regular, structured arrangement in texture-rich areas. This favorable spatial consistency directly contributes to improved performance in the reconstruction task.

\begin{table}[!t]
\centering
\caption{Ablation study. We  report matching capability (AUC@5° on MegaDepth~\cite{li2018megadepth}) and tracking stability (average value of AKTL on KITTI~\cite{geiger2012we} sequence 01-03).}
\label{tab:ablation}
\resizebox{0.76\columnwidth}{!}{
\begin{tabular}{lcc}
\toprule
Variant & AUC@5° $\uparrow$ & AKTL $\uparrow$  \\
\midrule
\textbf{TraqPoint-Full (Ours)} & \textbf{55.8} & \textbf{6.6} \\
\midrule
\multicolumn{3}{l}{\textit{1. RL Paradigm}} \\
\quad TraqPoint -- Pairwise & 53.3 & 4.3 \\
\quad Match Reward -- Pairwise & 49.7 & 2.8 \\

\midrule
\multicolumn{3}{l}{\textit{2. Reward Function Design}} \\
\quad w/o Ranking Reward & 52.6 & 4.0 \\
\quad w/o Distinctiveness Reward & 54.6 & 5.9 \\
\quad w/o RL learning & 52.0 & 3.8 \\
\midrule
\multicolumn{3}{l}{\textit{3. Backbone Architecture}} \\
\quad DINOv2-ViT & 54.8 & 6.4 \\
\quad ResNet-50  & 54.5 & 6.1 \\
\bottomrule
\end{tabular}
}
\vspace{-1.1 em}
\end{table}
\vspace{-0.3 em}
\subsection{Ablation}
\label{sec:ablation}
\vspace{-0.3 em}
We report two  metrics: AUC@5 on the MegaDepth dataset~\cite{li2018megadepth}, measuring matching capability. The AKTL metric on KITTI sequences 01–03~\cite{geiger2012we}, directly quantifying keypoint consistency. 
Results are shown in Tab.~\ref{tab:ablation}: 

\smallskip\noindent \textbf{1) Sequential vs. Pairwise Reinforcement Learning.} Two variants are constructed: (i) a two-frame simplified version, where rewards are computed solely on single image pairs to back propagate and optimize the policy network; (ii) using a basic matching reward (award keypoints that form nearest-neighbor matches with projected points), consistent with pairwise matching reward designs in most RL frameworks~\cite{tyszkiewicz2020disk, kunzel2025ripe}. Results show sequentialization improves AUC@5 by 2.5 and AKTL by 2.3. Additionally, our custom reward outperforms the basic match reward.

\smallskip\noindent \textbf{2) Reward Function Design.} We ablate our two reward components against a baseline of RDD-based supervised training (without RL). Results show that the ranking reward yields substantial performance gains by enforcing keypoint repeatability. The distinctiveness reward further elevates keypoint distinctiveness, effectively mitigating false matches. The supervised baseline exhibits marked performance degradation, underscoring the indispensable role of our RL framework in optimizing long-term trackability.

\smallskip\noindent \textbf{3) Backbone Architecture Analysis.}
Existing work confirms pre-trained backbones boost descriptor robustness~\cite{edstedt2024dedodev2, jiang2024Omniglue, chen2025rdd}. 
We evaluate two representatives: ResNet-50~\cite{he2016deep} and DINOv2-ViT~\cite{oquabdinov2}, where DINOv2-ViT only shows marginal gains over ResNet-50—we attribute this to its lack of multi-scale feature representation capabilities. Thus, we adopt DINOv3-ConvNeXt in experiments, as it possesses both multi-scale features and strong semantic representation abilities.
Notably, our sequence-aware RL yields consistent gains across all backbones, indicating the paradigm complements (not merely relies on) descriptor quality.

\begin{figure}[t]
    \centering
\includegraphics[width=0.48\textwidth]{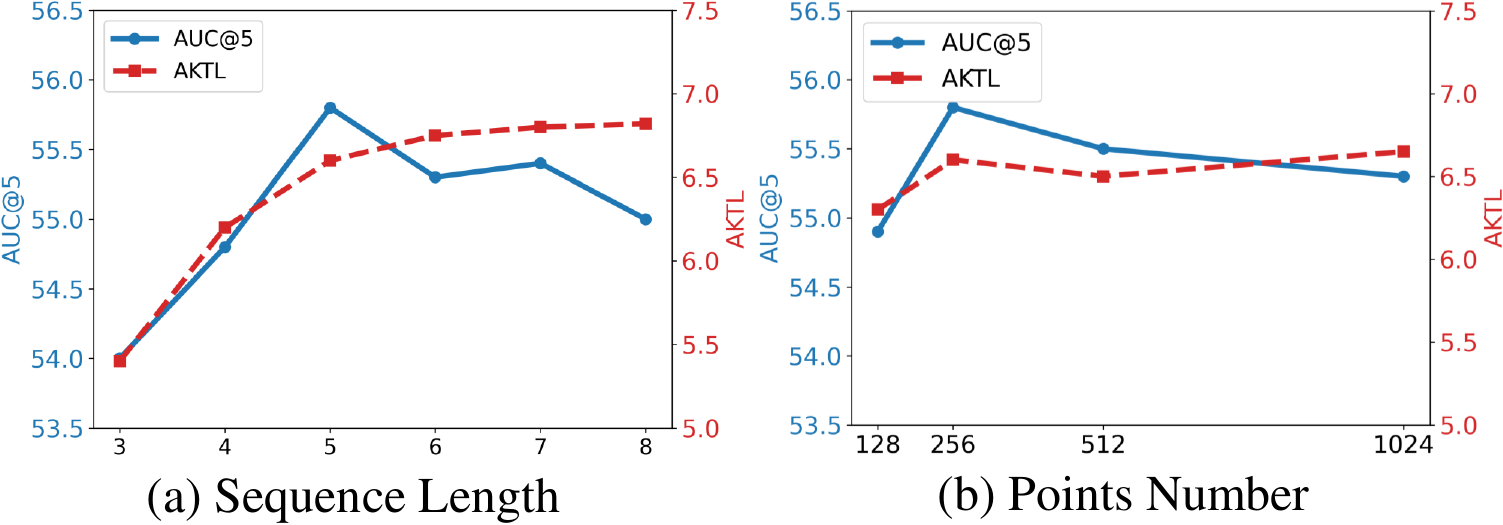}
    \vspace{-1.3 em}
    \caption{
 Ablation study on sequence length and the number of sampled keypoints.
 }
\label{fig:ablation}
\vspace{-1.3 em}
\end{figure}
\smallskip\noindent \textbf{4) Hyperparameter Sensitivity.}
We investigate two key hyperparameters: training sequence length and number of sampled keypoints. For  sequence length, increasing it enhances reward stability, but an excessively long sequence diminishes the average effective information. 
As shown in Fig.~\ref{fig:ablation}, the optimal parameters are a training sequence length of 5 and 256 sampled keypoints.

\vspace{-0.6 em}
\section{Conclusion}
\vspace{-0.4 em}
In this paper, we present a novel sequence-aware paradigm for learning local keypoints. We identify a critical gap in previous methods—the reliance on  pairwise matching objective that optimizes for short-term matchability rather than long-term consistency. To address this gap, we reformulate keypoint detection as a sequential decision-making problem and introduce a reinforcement learning framework trained with a novel composite reward that jointly enhances multi-view consistency and distinctiveness. Extensive experiments demonstrate that our approach outperforms existing methods across both pairwise and sequence-based downstream tasks. Thus, our work offers a novel research perspective for boosting the long-term stability of keypoints in sequential vision systems.

\noindent \textbf{Acknowledgment.}
This work was supported partially by the National Key Research and Development Program of China (2023YFC2705700), NSFC 62222112, 62176186, and the NSF of Hubei Province of China (2024AFA107).
\newpage
{
    \small
    \bibliographystyle{ieeenat_fullname}
    \bibliography{main}
}



\end{document}